\documentclass[letterpaper, 10 pt, conference]{ieeeconf}  
\pdfoutput=1
\usepackage{graphicx}
\usepackage{subfig}
\usepackage{cite}
\usepackage[table]{xcolor}
\usepackage{colortbl}
\usepackage{pgfplotstable}
\usepackage{numprint}
\usepackage{pgfplots}
\IEEEoverridecommandlockouts                              

\title{\LARGE \bf
A Decision-theoretic Approach to Detection-based Target Search with a UAV}

\author{Aayush Gupta$^{1}$, Daniel Bessonov$^{2}$, Patrick Li$^{3}$
\thanks{*The drone used in this project was funded by Micron Foundation through Saratoga High School IoT Club. This work was presented at IEEE IROS 2017.}
\thanks{$^{1}$Saratoga High School, 20300 Herriman Ave, Saratoga, CA 95070
		{\tt\small aayushgupta05@gmail.com}}%
\thanks{$^{2}$Saratoga High School, 20300 Herriman Ave, Saratoga, CA 95070
		{\tt\small danielbess16@gmail.com}}%
\thanks{$^{3}$Saratoga High School, 20300 Herriman Ave, Saratoga, CA 95070
		{\tt\small patrick8289@gmail.com}}%
}

\pgfplotstableset{
    /color cells/min/.initial=0,
    /color cells/max/.initial=1000,
    /color cells/textcolor/.initial=,
    color cells/.code={%
        \pgfqkeys{/color cells}{#1}%
        \pgfkeysalso{%
            postproc cell content/.code={%
                \begingroup
                \pgfkeysgetvalue{/pgfplots/table/@preprocessed cell content}\value
\ifx\value\empty
\endgroup
\else
                \pgfmathfloatparsenumber{\value}%
                \pgfmathfloattofixed{\pgfmathresult}%
                \let\value=\pgfmathresult
                \pgfplotscolormapaccess
                    [\pgfkeysvalueof{/color cells/min}:\pgfkeysvalueof{/color cells/max}]%
                    {\value}%
                    {\pgfkeysvalueof{/pgfplots/colormap name}}%
                \pgfkeysgetvalue{/pgfplots/table/@cell content}\typesetvalue
                \pgfkeysgetvalue{/color cells/textcolor}\textcolorvalue
                \toks0=\expandafter{\typesetvalue}%
                \xdef\temp{%
                    \noexpand\pgfkeysalso{%
                        @cell content={%
                            \noexpand\cellcolor[rgb]{\pgfmathresult}%
                            \noexpand\definecolor{mapped color}{rgb}{\pgfmathresult}%
                            \ifx\textcolorvalue\empty
                            \else
                                \noexpand\color{\textcolorvalue}%
                            \fi
                            \the\toks0 %
                        }%
                    }%
                }%
                \endgroup
                \temp
\fi
            }%
        }%
    }
}

\begin{document}

\maketitle
\thispagestyle{empty}
\pagestyle{empty}

\begin{abstract}

Search and rescue missions and surveillance require finding targets in a large area. These tasks often use unmanned aerial vehicles (UAVs) with cameras to detect and move towards a target. However, common UAV approaches make two simplifying assumptions. First, they assume that observations made from different heights are deterministically correct. In practice, observations are noisy, with the noise increasing as the height used for observations increases. Second, they assume that a motion command executes correctly, which may not happen due to wind and other environmental factors. To address these, we propose a sequential algorithm that determines actions in real time based on observations, using partially observable Markov decision processes (POMDPs). Our formulation handles both observations and motion uncertainty and errors. We run offline simulations and learn a policy. This policy is run on a UAV to find the target efficiently. We employ a novel compact formulation to represent the coordinates of the drone relative to the target coordinates. Our POMDP policy finds the target up to 3.4 times faster when compared to a heuristic policy.

\end{abstract}

\section{Introduction}

Unmanned aerial vehicles (UAVs) are an indispensable tool for search and rescue and can help find targets in critical, time sensitive missions \cite{search_rescue1}. However, these UAVs often fail to find the target when the uncertainty in the environment is high \cite{waharte_9}. 
The reason is that actions performed in real environments have stochastic outcomes and the UAV does not know its position and it cannot execute motions with certainty. 

\begin{figure}[thpb]
      \centering
      \includegraphics[width=0.5\columnwidth]{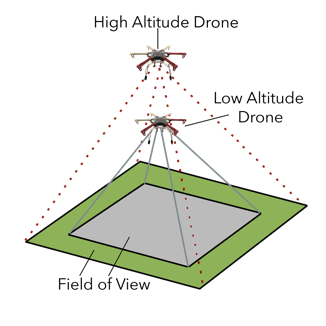}
      \caption{The lower the altitude, the higher the confidence and smaller the field of view of observations being made.}
      \label{f1}
   \end{figure}

When a drone issues a motion command in a certain direction, it does not always execute correctly. In addition, as the drone moves away from the ground, the field of view increases and the accuracy of the observation diminishes \cite{goodrich_4,waharte_8}. As shown in Figure \ref{f1}, as the drone moves down, the accuracy of the observation increases with the smaller field of view. This trade off between observation accuracy and altitude has been well studied. In \cite{goodrich_4}, the authors discuss the trade-off between the range of visibility and observation accuracy at different altitudes for a UAV. They provide images to a human operator, so that he can move the UAV remotely. 
  
\begin{figure}
\subfloat[MDP]{\includegraphics[width = 0.5\columnwidth]{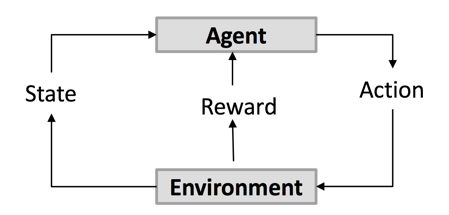}} 
\subfloat[POMDP]{\includegraphics[width = 0.5\columnwidth]{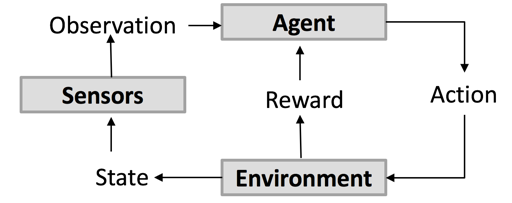}}
\caption{Left figure shows a MDP. Agent takes an action which updates its state. Infrequently agent may get a reward from the environment. The right figure shows a partially observable Markov decision process (POMDP) which is like a MDP except that the agent has a noisy access to its state through its sensors.}
\label{f2}
\end{figure}

Problems such as these have been well studied in a branch of AI called Reinforcement Learning. We can mathematically formulate this as a sequential decision making problem, where an agent has to make decisions at every time step using the knowledge of the world with uncertain action outcomes. The reward for acting correctly may not be reaped until many time steps later. A common assumption is that environment is observable, meaning that the agent knows where it is at all times. The outcome of each action is stochastic and can be captured using a transition model $T(s'| s,a)$, which gives the probability of transitioning to the next state $s'$ given that action $a$ is performed in state $s$. Transitions are typically assumed Markovian, such that this probability depends only on current state $s$ and not previous states. In addition there is a reward function $R(s)$ which encodes the utility of being in a certain state $s$. This formulation is called a Markov decision process (MDP), and consists of states, actions, transition model and reward function. The solution of an MDP is called a policy $\pi(s)$ and gives the action recommended for state $s$ \cite{russell_6}.

A real environment is typically not fully observable due to occlusions and sensor errors. The representation of such an environment is called a partially observable Markov decision process (POMDP). The agent does not know what state it is in so it cannot execute the action recommended by the optimal policy for that state. States are not observed directly, and are probabilistically inferred through noisy observations. As shown in Figure \ref{f2}, in an MDP the system performs an action based on its fully observable current state. In a POMDP however, the system performs an action based on the belief, which is a probabilistic distribution over possible states.

There has been past work utilizing POMDPs for UAVs. \cite{waharte_8} uses POMDPs for the drone’s decision making for search and rescue missions, with images as inputs. The authors state that with traditional grid platforms, the altitude must be kept constant. They partition the observation region into multiple grid cells, and have the sensor observe multiple grid cells depending on height. This grid-based representation of the environment is useful to describe the presence of the target in one grid cell\cite{partition1, partition2, partition3,partition4}. With this representation, authors of \cite{waharte_9} show it is simple to compute the probability of missed detections and false alarms as a function of height from field experiments. They also compare POMDP simulations to greedy and potential based strategies. The authors of \cite{chanel_1} discuss the modifications of UAVs in military operations by using POMDPs. They do not consider changes in height. The POMDP is solved online during the flight to take into account the current landscape. \cite{breckon_1,rudol_3,symington_7} supplement color recognition with thermal and/or radar data in order to improve efficiency and accuracy of detection. 

The problem with past work is the high dimensional state space $S$, which translates to a large belief space $B$. In this paper, first we introduce the concept of Relative State Space where the UAV does not keep track of absolute position of target, but only its position relative to the target. 
Second, we compare our POMDP derived policies to heuristic and random policies. Finally, we do a grid search to minimize search time by adjusting the parameters of the system such as transition probability, observation probability, and maximum reward.

\section{Our POMDP Formulation}
We use POMDPs, which are generalized versions of MDPs and model uncertainty in observations, in addition to movement. In our model we specify probabilities for observation accuracy as a function of height above ground and movement. We also specify a transition probability which is the probability of transitioning correctly to the next state after executing an action in the current state. Our POMDP framework is parameterized by the n-tuple: \{states, actions, observations, beliefs, transition function, reward function, discount factor\} \cite{waharte_9}.

\subsection{Reduced State Space}
The set of all possible states includes every position the drone could be in. States are parameterized by the n-tuple $S\{X, Y, Z, B, D\}$ where $X, Y, Z$ refers to position coordinates $\{1\leq X, Y, Z\leq N\}$, B is whether the drone is out of bounds, and D is whether the simulation is done. A single target state is designated as the reward state, such that any transition to the reward state delivers a reward. Our Reduced State Space representation models the coordinates of the drone relative to the reward coordinates. In detail, if the absolute coordinates of the drone are $(x_d, y_d, z_d)$, and the absolute coordinates of the reward are $(x_r, y_r, z_r)$, then the state is represented by $(x, y, z) = (x_d - x_r + c_x, y_d - y_r +c_y, z_d - z_r + c_z)$, where $c_x, c_y, c_z$ are constants that are used to make all the coordinates positive. This formulation is advantageous for several reasons: first, uncertainty in the reward location can be represented as an uncertainty in the current state instead; second, the current absolute location can be unknown because the location is only relative to the reward. Finally, this formulation leads to significant memory saving, rather than using $O(N ^ 5)$ space to store the coordinates of the reward and current state, only $O (N ^ 3) $ (with a constant factor of 4) space is used, where $N$ is the size of the X, Y, and Z dimensions. For example, when grid search size is 10, with absolute X coordinates there are 10 possible values for absolute state and 10 possible values of target giving rise to $N ^ 2=100$ possible x-coordinate values in the state. With relative coordinates, there are only $2N - 1=19$ possible x-coordinate values in the state. Y order computation is similar to X as described above, and for Z, the order is always N because the reward is known to be on level 1.

\subsection{Actions} 
There are 7 possible actions, involving motion or observation taking: east, west, north, south, ascend, descend, and look. In a non-stochastic world, north action moves the drone in the positive Y direction, west moves in the positive X direction, and ascend moves in positive Z direction. Look doesn’t move the current location. Instead, makes an observation through a sensor reading.

\subsection{Observations}
Unlike traditional systems, our observations are simply represented as a Boolean value representing whether or not the object can be seen in the frame, with a corresponding probability of accuracy. From the square of lowest height $Z = 1$, the drone can only see one square (its current square) with complete certainty. In general, at height $Z$, the drone can see $(2Z-1) ^ 2$ squares with accuracy of $a = \frac{1+{O_A} ^ {Z-1}}{2}$, where $O_A $ is the observation accuracy that varies from  $.7\leq O_A \leq 1.0$. With this equation, the accuracy is more than half and the information is generally more useful than no observation at all. Since observations are reported with observation accuracy that is available to the drone, the drone can calculate its precise Z coordinate. As a result, the problem becomes a mixed observability Markov decision process (MOMDP), a type of POMDP where the drone knows it’s precise Z coordinate but does not know X and Y coordinates. 

\subsection{Belief} 
Since the reward position relative to the drone is unknown, the initial state is unknown. Therefore, our initial belief of the current state is uniformly distributed over all states. Due to our relativistic formulation, we can either consider the system where the reward is known and the position of the drone is not, or the system where the position of the drone is exactly known and the reward is not. For simplicity, we consider the system where the reward is known and the position of the drone is not. As time progresses, the belief of the drone's current state moves with the perceived action taken, and is strengthened or weakened by the presence of positive or negative observations. Mathematically, if the probability of being in a certain state $S\{X, Y, Z, B, D\}$ is $P(s)$, the new probability after transitioning with action $A$ and observation $o$ is $P_{new} (s_0) = \sum\limits_{I = X - 1} ^ {X +1}\sum\limits_{J = Y - 1} ^ {Y +1}\sum\limits_{K = Z - 1} ^ {Z +1}\sum\limits_{B = 0}^ {1} P (S\{I, J, K, B, 0\})*T (s_0 | s, a)*O (o| s) $, where $T$ is the probability of transitioning to state $s_o$ after taking action $a$ in state $s$ and $O$ is the probability of observing $o$ in state $s$.

\subsection{Transition Function} 
When the drone attempts to execute an action, with the exception of look, it is possible that it executes a different action. We let the probability that the drone executes the correct action be $P$. The remaining probability is equally distributed over all of the 6 neighboring positions that are within bounds. As the state transitions between different coordinates, the belief updates accordingly. 

\subsection{Reward Function and Discount Factor} 
Every time the drone attempts to move out of bounds it gets a reward of $R_1 =-1$. Transitioning to the intended target has a reward $R_0 = 10$. Every time the drone executes an action, including look, one time step passes and the final reward’s value decreases proportionally to the discount factor $\gamma = .95$. Mathematically, the final reward value becomes $R = R_0 \gamma ^ T $, where $T$ is the number of time steps passed.

\section{Experiments}
We use the Julia POMDP library from Stanford Intelligent Systems Laboratory (SISL) for our simulations \cite{julia_10}. We use the SARSOP solver to compute the optimal POMDP policy. SARSOP is an approximate POMDP algorithm originally proposed by Kurniawati et. al in 2008 \cite{kurniawati_5}. Besides POMDPs, we coded two baseline policies for comparison:

\subsection{Heuristic Policy}
With this policy, the drone ascends till it sees the reward the first time, and every time it sees the reward thereafter, it descends. When it stops seeing the reward, it checks the 8 neighboring squares at the same level until it sees a reward again. When it oberserves a reward, it continues the descent. To maintain a maximum amount of information, it looks at every alternate turn. This policy models the thought process of a human operator.

\subsection{Random policy}
With a random policy, the drone has a 10\% chance of descending, otherwise it moves randomly on the same level.


 \begin{figure}[thpb]
      \centering
      \includegraphics[width=\columnwidth]{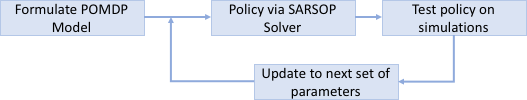}
      \caption{Schematic showing workflow for POMDP formulation, optimal policy determination and simulations.}
      \label{f4}
   \end{figure}

\subsection{Experimental Setup}
Figure \ref{f4} shows the workflow to optimize the POMDP policy through simulations. We use a 7x7x7 environment grid in our simulation. We vary the parameters of the model as follows: Reward is varied by an order of magnitude between 10 and 1000, while transition probability is varied between 0.76 to 1.0 with a spacing of 0.06. Similarly, observation probability is varied between 0.76 to 1.0 with a spacing of 0.06. Once we define our model, we run various mathematical solvers to heuristically determine near optimal policies through a SARSOP solver. 

We run each policy solver for 2 hours and optimize a policy which maps beliefs to actions. We then test this policy on 10,000 simulations with random starting states. The average reward from these runs is reported in the results section. We load this optimal policy onto the drone and query it during the flight to return the optimal direction of movement. 

\subsection{Drone Implementation}

\begin{figure}[tb]
      \centering
      \includegraphics[width=0.8\columnwidth]{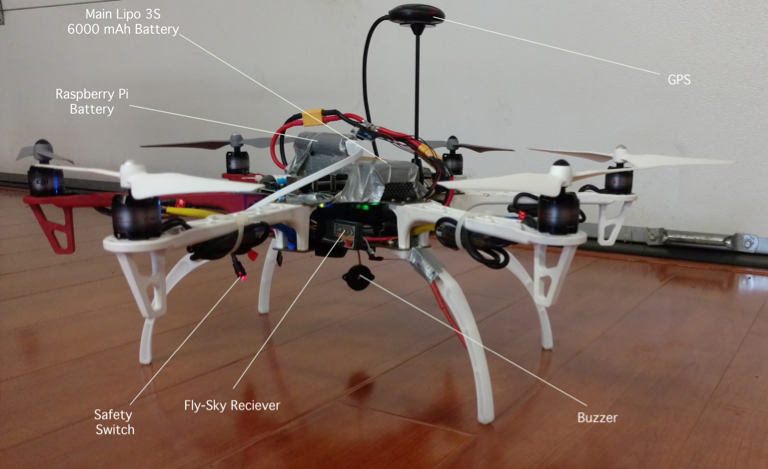}
      \caption{Our DJI F550 drone with labelled parts.}
      \label{f5}
   \end{figure}

\begin{figure}[tb]
      \centering
      \includegraphics[width=0.8\columnwidth]{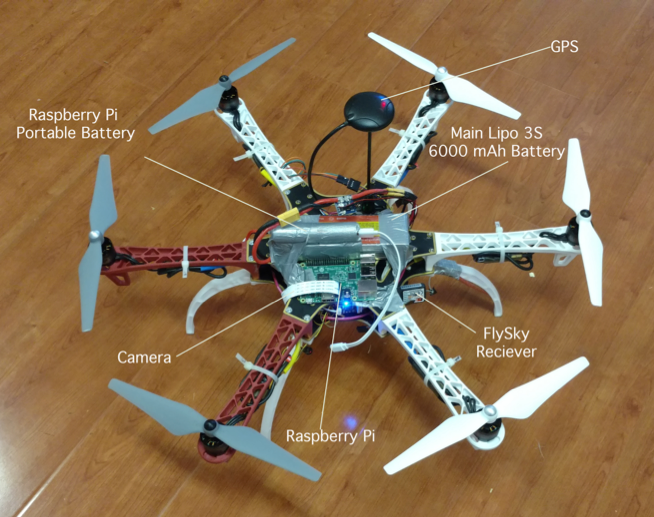}
      \caption{Top view of our DJI F550 drone with labelled parts.}
      \label{f6}
   \end{figure}

Our work uses the DJI F550, a hexacopter fitted with a 3DR Pixhawk Flight Control System and Raspberry Pi, connected to a high definition (1080p) camera. Our setup has the advantage of being programmable through the Raspberry Pi, as well as being controlled by a remote transmitter. Our drone is powered by a 3S lipo-cell battery, and a backup portable charger for the Raspberry Pi. It can carry upto 4 pounds including its initial hardware, with a flight time of approximately 15 minutes \cite{dji_11}. Our drone is shown in Figures \ref{f5} and \ref{f6}. It is equipped with a GPS and buzzer to notify the operator of any possible errors prior or during flight. We use a FlySky FS-T6 remote transmitter and AT9 receiver to begin the autonomous flight, as well as to maintain a manual override.                                                                                                                                                                                                                                                 

The Raspberry Pi 3 Model B is used to control all aspects of autonomous flight. Once the mission is started, a Python control script launches the UAV into a vertical ascent. For the current implementation, drone observations are hard-coded and read from a text file, as opposed to processing sensor data. This text file is parsed by a an update script written in Julia. This script processes the observation, and in turn determines and writes an action to the text file. The action is then executed by the drone, after which the process repeats until the reward is ultimately found.

\section{Results}
In this section, we first report on the actions choosen by the POMDP policy. We next compare our POMDP policy to a heuristic and random policy. Finally, we show the belief for a sample run of the POMDP policy in simulation and on the drone.

\begin{table}[!tb]
\caption{Proportion of time spent looking by the POMDP policy as a function of observation accuracy for transition probability = 0.88. As observations get more accurate, the system relies more on them.}
\label{look1}
\centering
\npdecimalsign{.}
\nprounddigits{2}
\npdigits{2}{2}
\begin{tabular}{|l|n{2}{2}|n{2}{2}|n{2}{2}|n{2}{2}|n{2}{2}|}
\hline
Observation &&&&&\\
Accuracy &0.76 &  0.82 &  0.88 & 0.94 &   1.0 \\
\hline
Look Proportion&0.00& 0.03& 0.10& 0.18& 0.21\\
\hline
\end{tabular}
\end{table}
\npnoround

\begin{table}[!tb]
\caption{Proportion of time spent looking by the POMDP policy as a function of transition probability for observation accuracy = .88. Generally, as movement gets more accurate, the system relies less on observation.}
\label{look2}
\centering
\npdecimalsign{.}
\nprounddigits{2}
\npdigits{2}{2}
\begin{tabular}{|l|n{2}{2}|n{2}{2}|n{2}{2}|n{2}{2}|n{2}{2}|}
\hline
Transition &&&&&\\
Probability &0.76 &  0.82 &  0.88 & 0.94 &   1.0 \\
\hline
Look Proportion&0.11& 0.11& 0.10& 0.09& 0.07\\
\hline
\end{tabular}
\end{table}
\npnoround

\begin{table}[!tb]
\caption{Average reward and time taken by POMDP, heuristic, and random policies over 1,000 simulations, varying observation probability and transition probability for R = 10.}
\label{t3}
\centering
\begin{tabular}{|l|c|c|}
\hline
Policy &  Average Reward & Average Time Steps \\
\hline
POMDP & 3.41 & 25.202 \\

Heuristic & 1.317 & 85.26 \\

Random & -2.112 & 142.05\\

\hline
\end{tabular}
\end{table}

\subsection{Comparison of POMDP to heuristic and random policy}
We use a 7x7x7 environment grid in our simulations. We vary transition probability and observation accuracy between 0.76 and 1. We compare the POMDP policy to heuristic and random policies for different values of observation accuracy and transition probability. Table \ref{t3} shows average reward and time taken by the policies assuming that each step takes 1 second. Note that it is not possible to have an exhaustive manual policy due to motion uncertainty. The average POMDP reaches a reward 3.4 times faster than the heuristic policy, and 5.7 times faster than the random policy.

  \begin{figure}[tb]
      \centering
      \includegraphics[width=3.7in]{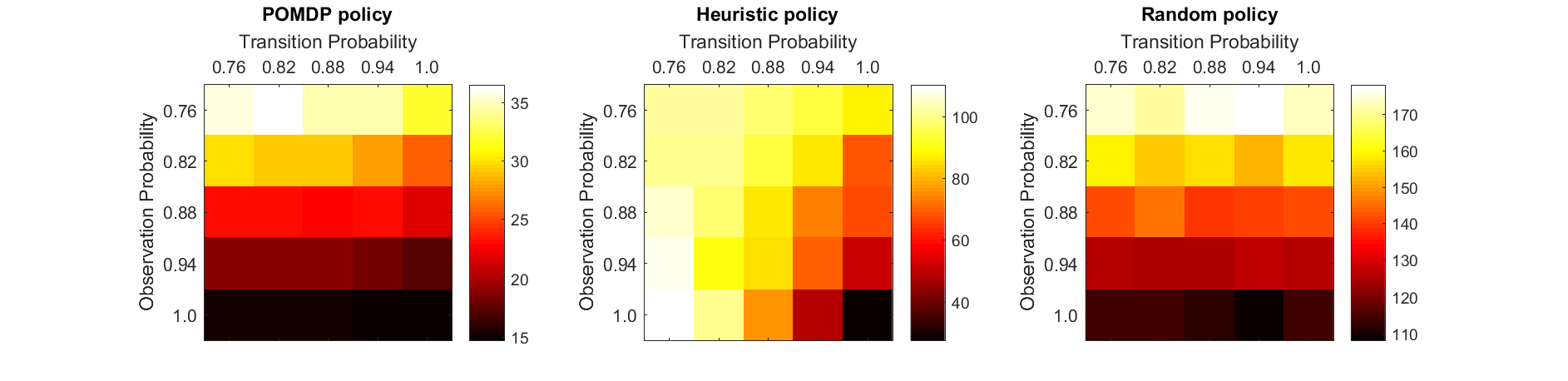}
      \caption{Heatmap showing time taken for POMDP, random and heuristic policies with R=10 with variation in transition probability on Y axis and observation probability on X axis. Note that the colormaps for each heatmap has a different range.}
      \label{f11}
   \end{figure}

\begin{figure}[tb]
\subfloat[Transition Accuracy Varying]{\includegraphics[width = 0.5\columnwidth]{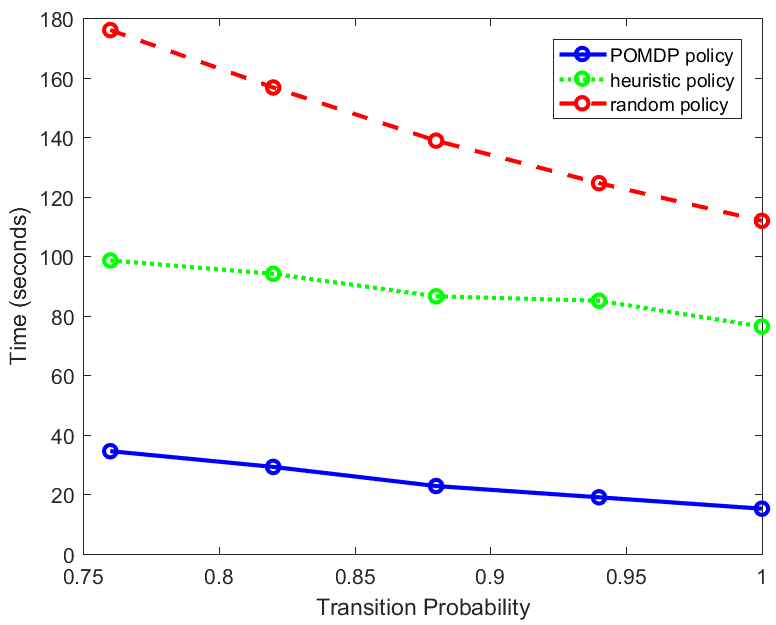}} 
\subfloat[Observation Accuracy Varying]{\includegraphics[width = 0.5\columnwidth]{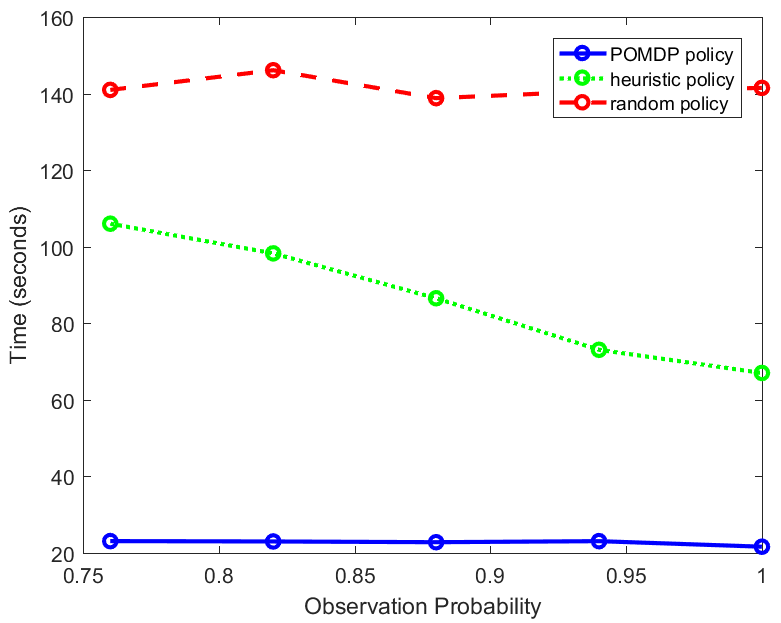}}
\caption{Both figures show time taken for POMDP, random and heuristic policies with R=10. The left figure varies transition probability while observation probability is 0.88. The right figure varies observation probability while transition probability is 0.88.}
\label{f12}
\end{figure}

In Figure \ref{f11}, we note that POMDP policy consistently takes less time. All the policies take less time when observation and transition probabilities are high as the environment is more certain, shown by the darker squares.
In Figure \ref{f12}(a) as transitions become more accurate, time to reward decreases for all policies as expected. In Figure \ref{f12}(b), we can see the POMDP policy is robust regarding uncertainties in observation and transitions, because time to destination remains similar as observation accuracy decreases, unlike the heuristic and random policies. We run 1,000 simulations for each policy and set of parameters to obtain accurate values.

\begin{figure}[!tbhp]
\centering
\subfloat[Step 1]{\includegraphics[width = 1.15in]{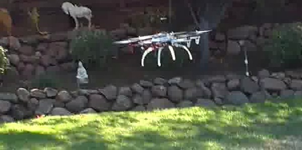}} 
\subfloat[Step 2]{\includegraphics[width = 1.15in]{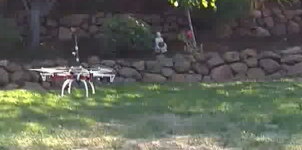}}
\subfloat[Step 3]{\includegraphics[width = 1.15in]{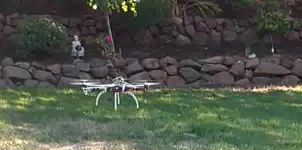}} \\
\subfloat[Step 4]{\includegraphics[width = 1.15in]{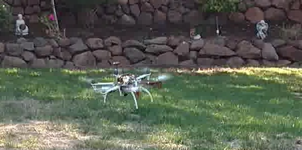}} 
\subfloat[Step 5]{\includegraphics[width = 1.15in]{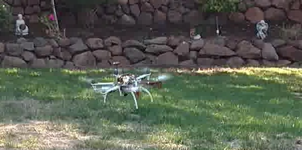}}
\subfloat[Step 6]{\includegraphics[width = 1.15in]{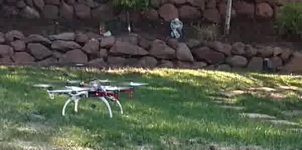}} \\
\subfloat[Step 7]{\includegraphics[width = 1.15in]{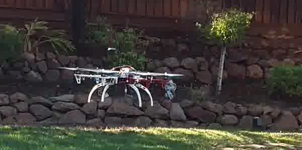}} 
\subfloat[Step 8]{\includegraphics[width = 1.15in]{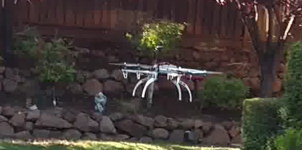}}
\subfloat[Step 9]{\includegraphics[width = 1.15in]{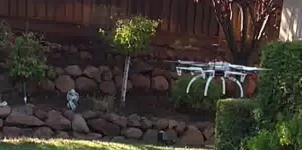}}\\
\caption{Steps 1-9. The drone follows the simulated POMDP policy under onboard software control moving to the left in Step 2, and then moving right in Step 3. This is followed by moving towards the user in Steps 4­-6. Finally, the drone moves to the right in Steps 7­-9.}
\label{f10}
\end{figure}

\begin{figure*}[!tbhp]
\centering
\subfloat[Step 1]{\includegraphics[width = 1.3in]{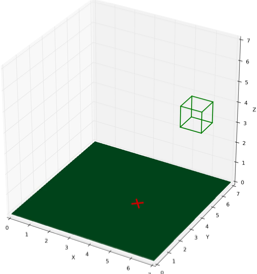}} 
\subfloat[Step 2]{\includegraphics[width = 1.3in]{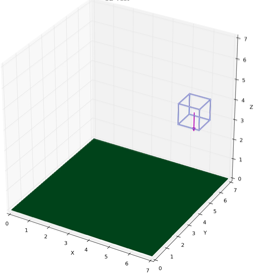}}
\subfloat[Step 3]{\includegraphics[width = 1.3in]{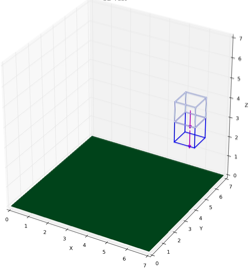}}
\subfloat[Step 4]{\includegraphics[width = 1.3in]{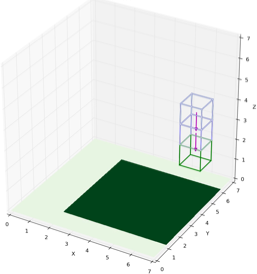}} \\ 
\subfloat[Step 5]{\includegraphics[width = 1.3in]{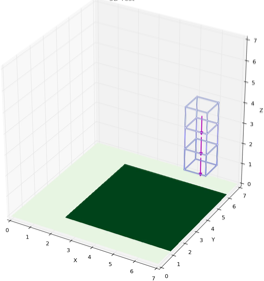}}
\subfloat[Step 6]{\includegraphics[width = 1.3in]{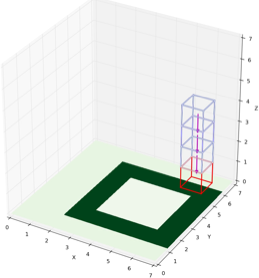}}
\subfloat[Step 7]{\includegraphics[width = 1.3in]{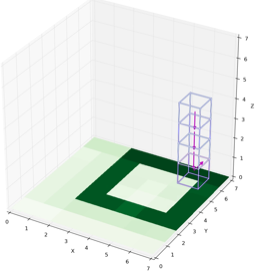}} 
\subfloat[Step 8]{\includegraphics[width = 1.3in]{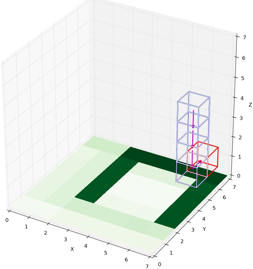}}\\ 
\caption{Showing steps 1-8 of a 16 step run.  In the first step (top left), the reward state (5,4) is shown with a red cross. The drone looks from (7,5,5) and finds the person from level 5 in the 11x11 grid on the ground centered on the UAV. It then decends from (7,5,5) to (7,5,4). In third step, the drone descends. In the 4th step at (7,5,3) it looks and finds a person, hence the belief is darker in the 5x5 grid around reward position (5,4). Note that the displayed belief can be intuitively understood as belief over UAV position. At (7,5,3), the UAV descends in top left quadrant. In step 6, the UAV is at (7,5,2) and looks down, not finding the person. Hence there is a light 3x3 region around (5,4), representing low belief in that region.}
\label{f8}
\end{figure*}

\begin{figure*}[!tbhp]
\centering
\subfloat[Step 9]{\includegraphics[width = 1.3in]{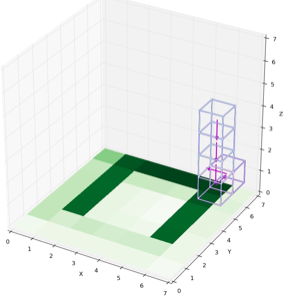}} 
\subfloat[Step 10]{\includegraphics[width = 1.3in]{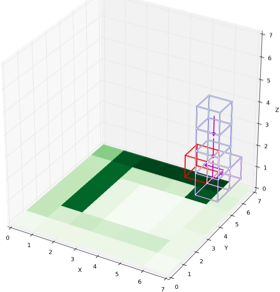}}
\subfloat[Step 11]{\includegraphics[width = 1.3in]{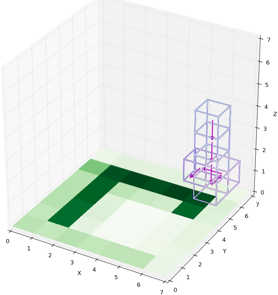}}
\subfloat[Step 12]{\includegraphics[width = 1.3in]{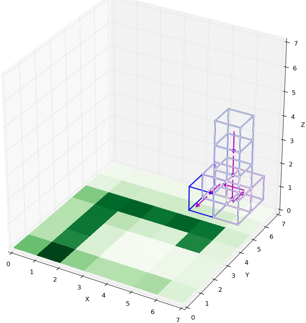}} \\ 
\subfloat[Step 13]{\includegraphics[width = 1.3in]{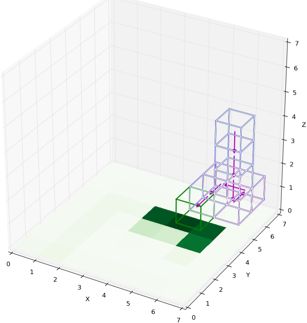}}
\subfloat[Step 14]{\includegraphics[width = 1.3in]{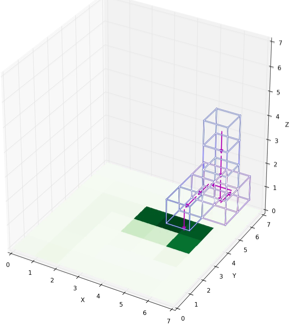}}
\subfloat[Step 15]{\includegraphics[width = 1.3in]{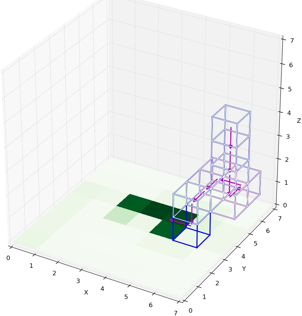}} 
\subfloat[Step 16]{\includegraphics[width = 1.3in]{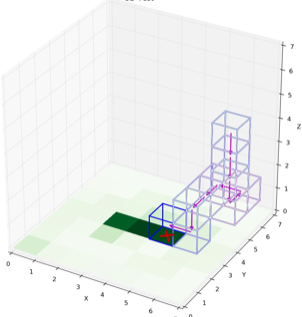}}\\ 
\caption{Steps 9-16. As simulation keeps moving forward the belief evolves, we notice that at step 13, the look action returns positive, increasing the belief in 3x3 grid around the position (5, 4). Rest of the grids get much lighter. At step 16, the system gets to reward state (5, 4), and the darkness of belief matches the true position. The final reward is shown with a red cross.}
\label{f9}
\end{figure*}

\subsection{Sample POMDP policy run in simulation and on drone}
In Figures \ref{f8} and \ref{f9}, we show an example run of the POMDP policy and walk through the example. A green cube outline signifies that the target was seen during a look action, a red cube outline signifies the target was not seen during a look action, and a blue cube signifies positions with non look actions. A cube outline with center $(X -.5, Y -.5, Z -.5)$ is said to be at position $(X, Y, Z)$. On the ground, squares with center $(X -.5, Y -.5)$  are said to be at position $(X, Y)$. In addition, darker green squares signify a higher belief that the current location is directly above the XY coordinate. Figure \ref{f10} shows a sample run on the drone.

\section{Conclusions and Future Work}

A common problem in application of POMDPs is their computational complexity. We reduce variables by defining state relative to the UAV’s own position, not absolute X and Y coordinates. This leads to significant memory saving; rather than using $O(N ^ 5)$ space to store the coordinates of the reward and current state, only $O (N ^ 3)$ space is used, where $N$ is the size of the X, Y, and Z dimensions. We build a simulator to test our Reduced State Space formulation for UAV policies performing target detection for search and rescue and surveillance tasks. We show that our POMDP policy finds the target 3.4 times faster as compared to a heuristic policy. We show a sample run of the POMDP policy where we are able to visualize the belief of the location of the reward relative to the drone, and run the policy on a drone.

In the future, we would like to expand our work in two different directions. First, we would like to integrate the camera sensor or a boolean sensor on our drone and take actions based on real observations. Second, we would like to do more experiments with other POMDP solvers such as Partially Observable Monte Carlo Planning where we do not learn a policy beforehand, but run simulations in real time based on sensor data to determine the optimal action.

\section*{ACKNOWLEDGMENT}
We would like to thank Prof. Mykel J. Kochenderfer at Stanford Intelligent Systems Laboratory (SISL) for guidance and Louis Dressel for paper review.

\bibliographystyle{ieeetr}
\bibliography{mybib}

\end{document}